\documentclass[11pt,a4paper]{article}
\usepackage[hyperref]{acl2018}
\usepackage{times}
\usepackage{latexsym}
\usepackage{url}
\usepackage{multirow}
\usepackage{epsfig}
\usepackage{amsmath}
\usepackage{amssymb}
\usepackage{subcaption}
\usepackage{graphicx}
\usepackage{bm}
\usepackage[group-separator={,},group-minimum-digits={3}]{siunitx}

\usepackage{CJKutf8}

\newcommand*{\affaddr}[1]{#1} % No op here. Customize it for different styles.
\newcommand*{\affmark}[1][*]{\textsuperscript{#1}}
\newcommand*{\email}[1]{\texttt{#1}}

\aclfinalcopy % Uncomment this line for the final submission
\def\aclpaperid{1546} %  Enter the acl Paper ID here

\title{Automatic Academic Paper Rating Based on Modularized Hierarchical Convolutional Neural Network}

\author{Pengcheng Yang\affmark[2], Xu Sun\affmark[1,2], Wei Li\affmark[1], Shuming Ma\affmark[1]\\
\affaddr{\affmark[1]MOE Key Lab of Computational Linguistics, School of EECS, Peking University}\\
\affaddr{\affmark[2]Deep Learning Lab, Beijing Institute of Big Data Research, Peking University}\\
\email{\{yang\_pc, xusun, liweitj47, shumingma\}@pku.edu.cn}\\
}

\date{}

\begin{document}
\maketitle
\begin{abstract}
As more and more academic papers are being submitted to conferences and journals, evaluating all these papers by professionals is time-consuming and can cause inequality due to the personal factors of the reviewers. In this paper, in order to assist professionals in evaluating academic papers, we propose a novel task: automatic academic paper rating (AAPR), which automatically determine whether to accept academic papers. We build a new dataset for this task and propose a novel modularized hierarchical convolutional neural network to achieve automatic academic paper rating. Evaluation results show that the proposed model outperforms the baselines by a large margin. The dataset and code are available at \url{https://github.com/lancopku/AAPR}
\end{abstract}

\section{Introduction}
Every year there are thousands of academic papers submitted to conferences and journals. Rating all these papers can be exhausting, and sometimes rating scores can be affected by the personal factors of the reviewers, leading to inequality problem. Therefore, there is a great need for rating academic papers automatically. In this paper, we explore how to automatically rate the academic papers based on their \LaTeX \  source file and meta information, which we call the task of automatic academic paper rating (AAPR). 

A task that is similar to the AAPR is automatic essay scoring (AES). AES has been studied for a long time. Project Essay Grade \citep{Page1966Grading,Page1968The} is one of the earliest attempts to solve the AES task by predicting the score using linear regression over expert crafted textual features. Much of the following work applied similar methods by using various classifiers with more sophisticated features including grammar, vocabulary and style \citep{Rudner2002Automated,Attali2006AUTOMATED}. These traditional methods can work almost as well as human raters. However, they all demand a large amount of feature engineering, which requires a lot of expertise. 

Recent studies turn to use deep neural networks, claiming that deep learning models can relieve the system from heavy feature engineering. \citet{Dimitrios2016Automatic} proposed to use long short term memory network \citep{Hochreiter1997Long} with a linear regression output layer to predict the score. They added a score prediction loss to the original C\&W embedding \citep{Collobert2008A, Collobert2011Natural}, so that the word embeddings are related to the quality of the essay. \citet{Taghipour2016A} also applied recurrent neural networks to process the essay, except that they put a convolutional layer ahead of the recurrent layer to extract local features. \citet{Dong2016Automatic} proposed to apply a two-layer convolutional neural network (CNN) to model the essay. The first layer is responsible for encoding the sentence and the second layer is to encode the whole essay. \citet{Dong2017Attention} further proposed to add attention mechanism to the pooling layer to automatically decide which part is more important in determining the quality of the essay.

Although there has been a lot of work dealing with AES task, researchers have not attempted the AAPR task. Different from the essay in language capability tests, academic papers are much longer with much more information, and the overall quality is affected by a variety of factors besides the writing. Therefore, we propose a model that considers the overall information of one academic paper, including the title, authors, abstract and the main content of the \LaTeX \ source file of the paper.

Our main contributions are listed as follows:
\begin{itemize}
\item We propose the task of automatically rating academic papers and build a new dataset for this task. 
\item We propose a modularized hierarchical convolutional neural network model that considers the overall information of the source paper. Experimental results show that the proposed method outperforms the baselines by a large margin. 
\end{itemize}

\section{Proposed Method}
\begin{figure}[tb]
	\centering
	\subcaptionbox{Modularized hierarchical convolutional neural network.}{\includegraphics[width=1.0\linewidth]{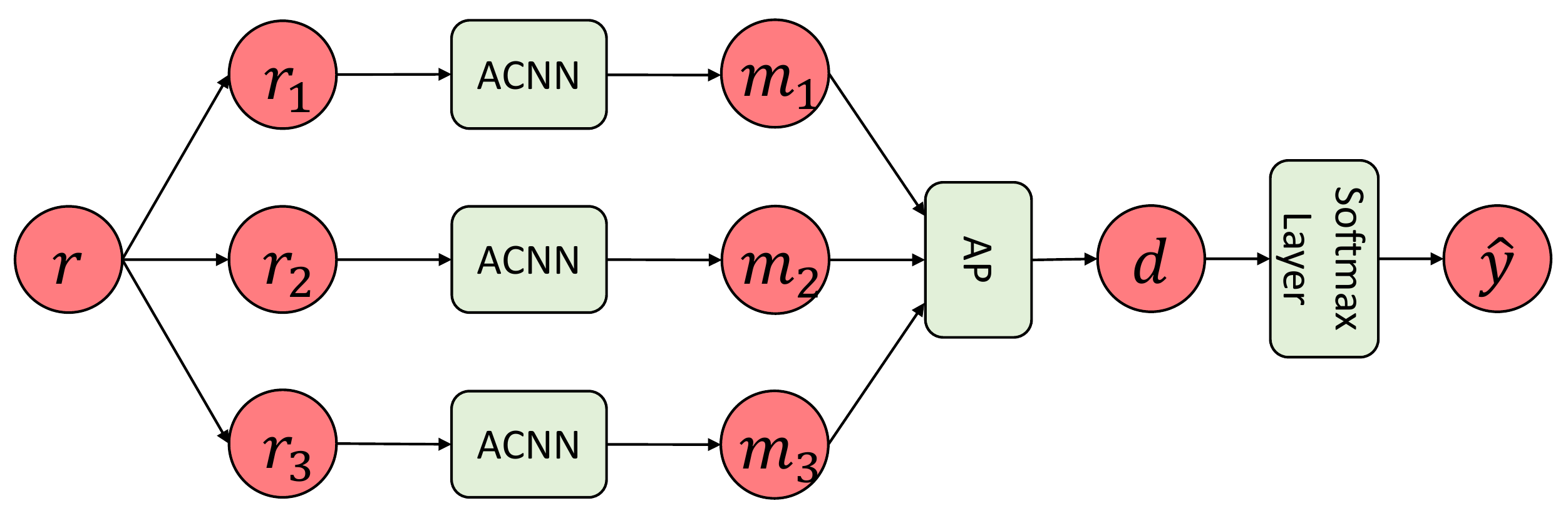}} 
    \subcaptionbox{Attention-based convolutional neural network.}{\includegraphics[width=1.0\linewidth]{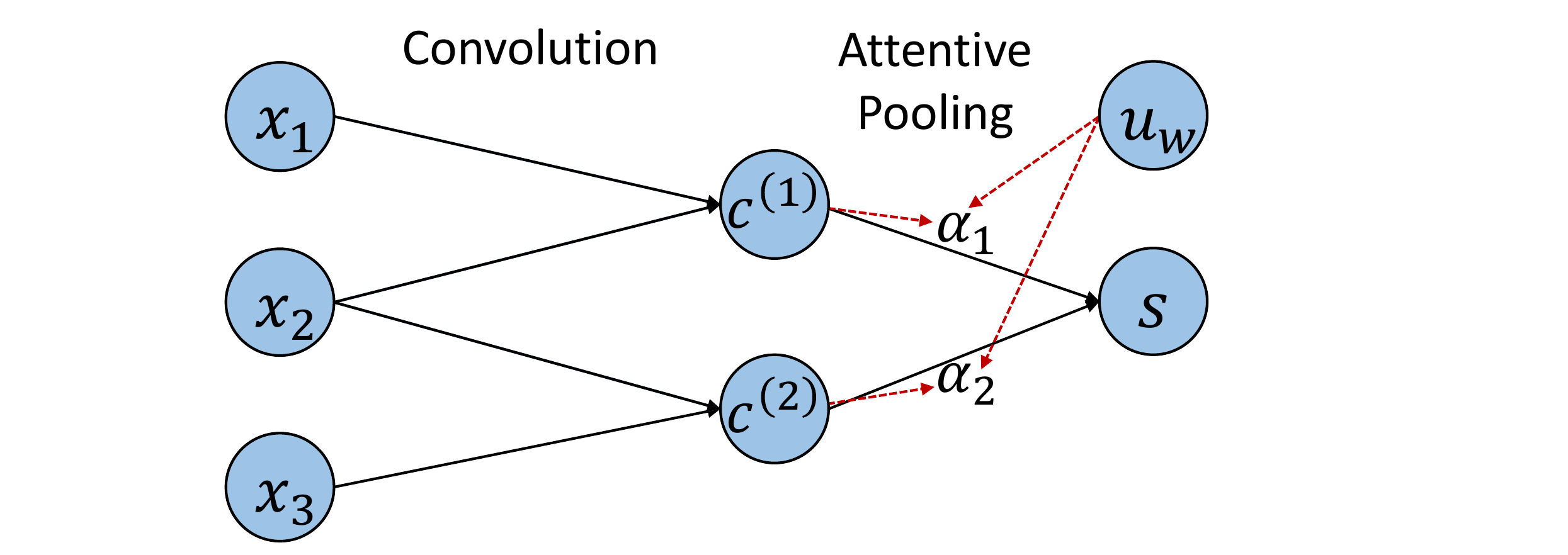}}
	\caption{The overview of our model. ACNN denotes attention-based CNN, whose basic structure is shown in (b). AP denotes attentive pooling.}\label{Fig:model_fig}
	\vspace{-0.05in}
\end{figure}

A source paper usually consists of several modules, such as \emph{abstract}\footnote{Italicized words represent modules of the source paper.}, \emph{title} and so on. There is also a hierarchical structure from word-level to sentence-level in each module. The structure information is likely to be helpful to make more accurate predictions. Besides, the model can be improved by considering the difference in contributions of various parts of the source paper. 
Based on this observation, we propose a modularized hierarchical CNN. An overview of our model is shown in Figure~\ref{Fig:model_fig}. We assume that a source paper has $l$ modules, with $m$ words and the filter size is $h$ (detailed explanations can be referred to Section~\ref{mhcnn} and Section~\ref{acnn}). $l, m$ and $h$ are set to be 3, 3, 2, respectively in the Figure~\ref{Fig:model_fig} for simplicity.

\subsection{Modularized Hierarchical CNN}\label{mhcnn}
Given a complete source paper $\bm{r}$, represented by a sequence of tokens, we first divide it into several modules $(\bm{r_1}, \bm{r_2}, \cdots, \bm{r_l})$ based on the general structure of the source paper (\emph{abstract, title, authors, introduction, related work, methods and conclusion}). For each module, the one-hot representation of the $i$-th word $\bm{w}_i$ is embedded to a dense vector $\bm{x}_i$ through an embedding matrix.
%$\bm{W} \in \mathbb{R}^{k \times |\mathcal{V}|}$, where $|\mathcal{V}|$ is the size of vocabulary. 
For the following modules (\emph{abstract, introduction, related work, methods, conclusion}), we use the attention-based CNN (illustrated in Section~\ref{acnn}) in word-level to get the representation $\bm{s}_i$ of the $i$-th sentence. Another attention-based CNN layer is applied to encode the sentence-level representations into the representation $\bm{m}_i$ of the $i$-th module.

There is only one sentence in the title of the source paper, so it is reasonable to get the module-level representation of \emph{title} only using attention-based CNN in word-level. Besides, the weighted average method is applied to obtain the module-level representation of \emph{authors} by Equation~\eqref{authors} because the authors are independent of each other.
\begin{equation}\label{authors}
\bm{m}_{authors} = \sum_{i=1}^A\gamma_i\bm{a}_i 
\end{equation}
where $\bm{\gamma} = (\gamma_1, \ldots, \gamma_A)^T$ is the weight parameter. $\bm{a}_i$ is the embedding vector of the $i$-th author in the source paper, which is randomly initialized and can be learned at the training stage. $A$ is the maximum length of the author sequence. 

Representations $\bm{m}_1, \bm{m}_2, \cdots, \bm{m}_l$ of all modules are aggregated to form the paper-level representation $\bm{d}$ of the source paper with an attentive pooling layer. A $softmax$ layer is used to take $\bm{d}$ as input and predict the probability of being accepted. At the training stage, the cross entropy loss function is optimized as objective function, which is widely used in various classification tasks.
\begin{equation}\label{softmax}
\hat{\bm{y}} = softmax(\bm{W}_d \bm{d} + \bm{b}_d)
\end{equation}

%AP denotes the attentive pooling. Red dotted lines represent the attention pooling.
% For simplicity, we assume $l, m$, and $h$ equal to 3, 3, 2, respectively.

\subsection{Details of Attention-Based CNN}\label{acnn}
Attention-based CNN consists of a convolution layer and an attentive pooling layer. The convolution layer is used to capture local features and attentive pooling layer can automatically decide the relative weights of words, sentences, and modules.

\textbf{Convolution layer:} 
A sequence of vectors of length $m$ is represented as the row concatenation of $m$ $k$-dimensional vectors: $ \bm{X} = [\bm{x}_1; \bm{x}_2; \cdots; \bm{x}_m]
$. A filter $\bm{W}_x\in \mathbb{R}^{h \times k}$ convolves with the window vectors at each position to generate a feature map $\bm{c} \in \mathbb{R}^{m-h+1}$. Each element $c_j$ of the feature map is calculated as follows:
\begin{equation}\label{concov}
c_j = f(\bm{W}_x \circ [\bm{x}_j:\bm{x}_{j+h-1}] + b_x)
\end{equation}
where $\circ$ is element-wise multiplication, $b_x \in \mathbb{R}$ is a bias term, and $f$ is a non-linear activation function. Here we choose $f$ to be ReLU \cite{relu}. $n$ different filters can be used to extract multiple feature maps $\bm{c}_1, \bm{c}_2, \cdots, \bm{c}_n$. We get new feature representations $\bm{C}\in \mathbb{R}^{(m-h+1)\times n}$ as the column concatenation of feature maps $\bm{C} = [\bm{c}_1, \bm{c}_2, \cdots, \bm{c}_n]$. The $i$-th row $\bm{c}^{(i)}$ of $\bm{C}$ is the new feature representation generated at position $i$.  

\textbf{Attentive pooling layer:} 
Given a sequence $\bm{c}^{(1)}, \bm{c}^{(2)}, \cdots, \bm{c}^{(q)}$, which are $q$ $n$-dimensional vectors, the attentive pooling is applied to aggregate the representations of the sequence by measuring the contribution of each vector to form the high-level representation $\bm{s}$ of the whole sequence. Formally, we have
\begin{align}
\bm{z}_i & = \tanh(\bm{W}_c \bm{c}^{(i)} + \bm{b}_c) \label{attent1.1} \\
\alpha_i & = \frac{\bm{z}_i^T\bm{u}_w}{\sum_{k}\exp(\bm{z}_k^T\bm{u}_w)} \label{attent1.2} \\
\bm{s} & = \sum_i\alpha_i \bm{z}_i \label{attent1.3}
\end{align} 
where $\bm{W}_c$ and $\bm{b}_c$ are weight matrix and bias vector, respectively. $\bm{u}_w$ is a randomly initialized vector, which can be learned at the training stage.

\section{Experiments}
In this section, we evaluate our model on the dataset we build for this task. We first introduce the dataset, evaluation metric, and experimental details. Then, we compare our model with baselines. Finally, we provide the analysis and the discussion of experimental results.

\subsection{Dataset}
\noindent\textbf{Arxiv Academic Paper Dataset:} As there is no existing dataset that can be used directly, we create a dataset by collecting data on academic papers in the field of artificial intelligence from the website\footnote{\url{https://arxiv.org/}}. The dataset consists of 19,218 academic papers. The information of each source paper consists of the venue which marks whether the paper is accepted, and the source \LaTeX \ file. We divide the dataset into training, validation, and test parts. The details are shown in Table \ref{tab_dataset_numbers}.

%To apply the proposed method on this dataset, we extract (title, authors, abstract, introduction, related work, methods, conclusion, venue) tuples from papers' source \LaTeX \ files and metadata. We use all these information except the venue in the tuple to predict whether the paper can be accepted. We divide the dataset into three parts, the statistical information of the three parts are shown in Table~\ref{tab_dataset_numbers}.

%Table~\ref{tab_average_num_words} reports the average number of words and sentences.
%\begin{table}[htb]
%	\centering
%	\normalsize
%	\begin{tabular}{p{2.8cm}lcc}
%		\hline
%		\multicolumn{1}{l}{\textbf{Section}} &
%        \multicolumn{1}{c}{\textbf{\# words}} &
%		\multicolumn{1}{c}{\textbf{\# sentences}}   \\ \hline
%		Title & 8.63 & 1\\
%       Authors & 4.67 & 1 \\
%        Abstract & 14.29 & 6.78\\
%        Introduction &  9.75 & 62.43 \\
%        Related work & 8.23 & 298.63 \\
%        Methods &  7.59 & 413.68 \\
%		Conclusion & 9.64 & 102.26\\ \hline
%	\end{tabular}
%	\caption{Average length of words and sentences. \# words represents the average number of words in a sentence. \# sentences represents the average number of sentences in a module.}\label{tab_average_num_words}
%\end{table}

\begin{table}[tb]
		\centering
		\begin{tabular}{ l  c  c  c }
		\hline
		\multicolumn{1}{l}{\textbf{Dataset}} &
		\multicolumn{1}{c}{\#\textbf{Total}} & 
		\multicolumn{1}{c}{\#\textbf{Positive}} &  
		\multicolumn{1}{c}{\#\textbf{Negative}}   \\ \hline
		Training set & \num{17218} & \num{8889} & \num{8329} \\ 
		Validation set & \num{1000} & \num{507} & \num{493} \\ 
		Test set & \num{1000} & \num{504} & \num{496} \\ \hline
		\end{tabular}
		\caption{Statistical information of Arxiv academic paper dataset. \textbf{Positive} and \textbf{Negative} denote whether the source paper is accepted.}
		\label{tab_dataset_numbers}
	\end{table}

\subsection{Experimental Details}
%Here we provide the details of experiment, including evaluation metrics, model parameters and training parameters.

We use accuracy as our evaluation metric instead of the F-score, precision, and recall because the positive and negative examples in our dataset are well balanced.

Since the author names are different from the common scientific words in the paper, we separately build up vocabulary for authors and text words of source papers with the size of \num{20000} and \num{50000}, respectively. 
%We limit the vocabulary to a certain number of most frequent words appeared in the training set and The size of text-vocabulary and author-vocabulary is .

We use the training strategies mentioned in \citet{cnn_trick} for CNN classifier to tune the hyper-parameters based on the accuracy on the validation set. The word or author embedding is randomly initialized and can be learned during training. The size of word embedding or author embedding is 128 and the batch size is 32. Adam optimizer~\cite{KingmaBa2014} is used to minimize cross entropy loss function. We apply dropout regularization~\cite{dropout} to avoid overfitting and clip the gradients~\cite{gradientclip} to the maximum norm of 5.0. 

During training, we train the model for a fixed number of epochs and monitor its performance on the validation set after every 50 updates. Once training is finished, we select the model with the highest accuracy on the validation set as our final model and evaluate its performance on the testing set.

\subsection{Baselines}
We compare our model with the following baselines:

\begin{itemize}
	\item \noindent\textbf{Randomly predict (RP):} We randomly decide whether the source paper can be accepted. In other words, the probability of acceptance of every source paper is always 0.5 using this strategy.
	\item \noindent\textbf{Traditional machine learning algorithms:} We use various machine learning classifiers to predict the labels based on the tf-idf features of the text. %The word embedding technique is not used.
	\item \noindent \textbf{Neural networks models:} We apply three representative neural network models: CNN~\cite{cnn}, LSTM~\cite{Hochreiter1997Long}, and C-LSTM~\cite{c-lstm}. We concatenate all modules of the source paper into a long text sequence as the input to the neural network models.
\end{itemize}

\subsection{Results}

\begin{table}[tb]
	\centering
	\begin{tabular}{l c l c}
		\hline
		\multicolumn{1}{l}{\textbf{Models}} &
        \multicolumn{1}{l}{\textbf{Accuracy}} &
        \multicolumn{1}{l}{\textbf{Models}} &
		\multicolumn{1}{c}{\textbf{Accuracy}} \\ \hline
        RP & 50.0\% & Logistic & 60.0\% \\
        CART & 58.6\% & KNN & 60.3\% \\
        MNB & 58.3\% & GNB & 58.5\% \\
		SVM & 61.6\% & AdaBoost & 58.9\% \\
        Bagging & 59.4\% & LSTM & 60.5\% \\
        %Forest & 60.5\% \\ \hline
        CNN & 61.3\% & C-LSTM & 60.8\% \\ \hline
		%CNN & 62.6\% \\
		%CNN + AP & 64.8\% \\ 
        %HCNN & 66.8\% \\
        \textbf{MHCNN} & \textbf{67.7\%} & &\\ \hline
	\end{tabular}
	\caption{Comparison between our proposed model and the baselines on the test set. Our proposed model is denoted as \textbf{MHCNN}.}\label{tab_part1_res}
\end{table}

In this subsection, we present the results of evaluation by comparing our proposed method with the baselines. Table~\ref{tab_part1_res} reports experimental results of various models. As is shown in Table~\ref{tab_part1_res}, the proposed MHCNN outperforms all the above mentioned baselines. The best baseline model SVM achieves the accuracy of 61.6\%, while the proposed model achieves the accuracy of 67.7\%. In addition, our MHCNN outperforms other representative deep-learning models by a large margin. For instance, the proposed MHCNN achieves an improvement of 6.4\% accuracy over the traditional CNN. This shows that our MHCNN can learn better representation by considering modularized hierarchical structure in the source paper. Our proposed MHCNN aims to divide a long text into several modules and using attention mechanism to aggregate the representations of each module to form a final high-level representation of a complete source paper. By incorporating knowledge of the structure of the source paper and automatically selecting the most informative words, the model is capable of making more accurate predictions.

\section{Analysis and Discussions}
Here we perform further analysis on the model and experiment results.

\subsection{Exploration on Internal Structure of the Model}

\begin{table}[t]
	\centering
	\begin{tabular}{@{}l@{}c@{}c@{}}
		\hline
		\multicolumn{1}{l}{\textbf{Models}} &
		\multicolumn{1}{c}{\textbf{Accuracy}} &
        \multicolumn{1}{c}{\textbf{Decline}} \\ \hline
        MHCNN & 67.7\% & $--$ \\ \hline
		w/o Attention & 66.8\%* & $\downarrow$0.9\% \\
		w/o Module & 61.3\%* & $\downarrow$6.4\% \\ \hline
	\end{tabular}
	\caption{Ablation Study. The symbol * indicates that the difference compared to MHCNN is significant with $p \leq 0.05$ under $t$-test.}\label{tab_part1_res2}
\end{table}

As is shown in Table~\ref{tab_part1_res}, our MHCNN model outperforms all baselines by a large margin. Compared with the basic CNN model, the proposed model has a modularized hierarchical structure and uses multiple attention mechanisms. In order to explore the impact of internal structure of the model, we remove the modularized hierarchical structure and attention mechanisms in turn. The performance is shown in Table~\ref{tab_part1_res2}. “w/o Attention” means that we still use modularized hierarchical structure while do not use any attention mechanism. “w/o Module” means that we do not use both attention mechanism and modularized hierarchical structure, which is the same as the CNN model in the baselines.

As is shown in Table~\ref{tab_part1_res2}, the accuracy of the model drops by 0.9\% when the attention mechanism is removed from the model. This shows that there are differences in the contribution of textual content. For instance, the \emph{abstract} of a source paper is more important than its \emph{title}. Attention mechanism can automatically decide the relative weights of modules, which makes  model predictions more accurate. However, the accuracy of the model drops by 6.4\% when we remove the modularized hierarchical structure, which is much larger than 0.9\%. It shows that the modularized hierarchical structure of the model is of great help to obtain better representations by incorporating knowledge of the structure of the source paper.

\subsection{The Impact of Modules of the Source Paper}
%The evaluation results show that the organizational structure of the source paper has a great impact on its acceptance probability. 
One interesting issue is which part of the source paper best determines whether it can be accepted. To explore this issue, we subtract each module from complete source papers in turn and observe the change in the performance of the model. The experimental result is shown in Table~\ref{tab_part2_res}. 

As is shown in Table~\ref{tab_part2_res}, the performance of the model shows different degrees of decline when we remove different modules of the source paper. This shows that there are differences in the contribution of different modules of the source paper to its acceptance, which further illustrates the reasonableness of our use of modularized hierarchical structure and attention mechanism. All the declines are significant with $p \leq 0.05$ under the $t$-test. 

When we remove \emph{authors} module, the accuracy drops by 3.1\%, which is the largest decline. This shows that the authors of the source paper largely determines whether it can be accepted. Obviously, a source paper written by a proficient scholar tends to be good work, which has a higher probability of being accepted. Except for \emph{authors}, the two most significant modules affecting the probability of being accepted are \emph{conclusions} and \emph{abstract}. Because they are the essence of the entire source paper, which can directly reflect the quality of the source paper. 
%In the main body of source papers, the conclusion occupies a crucial part while the methods has little effect. 
However, the \emph{methods} module of the source paper has little effect on the probability of being accepted according to Table~\ref{tab_part2_res}. The reason may be that the \emph{methods} of different source papers vary widely, which means that there exists high variance in this module. Therefore, our model may not do well in capturing a unified internal pattern to make prediction. The impact of the \emph{title} is the smallest and the accuracy of the model drops by only 1.1\% when \emph{title} is removed from the source paper. 
 
\begin{table}[t]
	\centering
	\begin{tabular}{@{}l@{}c@{}c@{}}
		\hline
		\multicolumn{1}{l}{\textbf{Contexts}} &
		\multicolumn{1}{c}{\textbf{Accuracy}} &
        \multicolumn{1}{c}{\textbf{Decline}} \\ \hline
        Full data & 67.7\% & $--$ \\ \hline
        w/o \emph{Title} & 66.6\%* & $\downarrow$1.1\% \\
		w/o \emph{Abstract} & 65.5\%* & $\downarrow$2.2\% \\
		w/o \emph{Authors} & 64.6\%* & $\downarrow$3.1\% \\
		w/o \emph{Introduction} &  65.7\%* & $\downarrow$2.0\% \\
		w/o \emph{Related work} &  66.0\%*  & $\downarrow$1.7\% \\
        w/o \emph{Methods} & 66.2\%* & $\downarrow$1.5\% \\
        w/o \emph{Conclusion} & 65.0\%* & $\downarrow$2.7\%  \\ \hline
	\end{tabular}
	\caption{Ablation Study. The symbol * indicates that the difference compared to full data is significant with $p \leq 0.05$ under $t$-test.}\label{tab_part2_res}
\end{table}

\section{Related Work}
The most relevant task for our work is automatic essay scoring (AES). There are two main types of methods for the AES task: traditional machine learning algorithms and neural network models.

Most traditional methods for the AES task use supervised learning algorithms, including classification \cite{larkey1998automatic, Rudner2002Automated, yannakoudakis2011new, chen2013automated}, regression \cite{Attali2006AUTOMATED, phandi2015flexible, zesch2015task} and so on. However, they all require lots of manual features, for instance, bag of words, spelling errors, or lengths, which can be time-consuming and requires a large amount of expertise.

In recent years, some neural network models have also been used for the AES task, which have achieved great success. \citet{Dimitrios2016Automatic} proposed to use the LSTM model with a linear regression output layer to predict the score. \citet{Taghipour2016A} applied the CNN model followed by a recurrent layer to extract local features and model sequence dependencies. A two-layer CNN model was proposed by \citet{Dong2016Automatic} to cover more high-level and abstract information. \citet{Dong2017Attention} further proposed to add attention mechanism to the pooling layer to automatically decide which part is more important in determining the quality of the essay. \citet{song2017discourse} proposed a multi-label neural sequence labeling approach for discourse mode identification and showed that features extracted by this method can further improve the AES task.

\nocite{ma2017cascading}
\nocite{wei2017minimal}
\nocite{meng2013unified}
\nocite{chen2018modeling}

\section{Conclusions}
In this paper, we propose the task of automatic academic paper rating (AAPR), which aims to automatically determine whether to accept academic papers. We propose a novel modularized hierarchical CNN for this task to make use of the structure of a source paper. Experimental results show that the proposed model outperforms various baselines by a large margin. In addition, we find that the conclusion and abstract parts have the most influence on whether the source paper can be accepted when setting aside the factor of authors.

\section{Acknowledgements}
This work is supported in part by National Natural Science Foundation of China (No. 61673028), National High Technology Research and Development Program of China (863 Program, No. 2015AA015404), and the National Thousand Young Talents Program. Xu Sun is the corresponding author of this paper.

\bibliography{acl2018}
\bibliographystyle{acl_natbib}

\end{document}